\documentclass[10pt]{article}

\newif\ifanon
\anonfalse

\usepackage[utf8]{inputenc}
\usepackage[T1]{fontenc}
\usepackage[letterpaper,margin=1.0in]{geometry}
\usepackage{mathptmx}
\usepackage{amsmath,amssymb,amsthm}
\usepackage{booktabs}
\usepackage{graphicx}
\usepackage{caption}
\usepackage{subcaption}
\usepackage{xcolor}
\usepackage{microtype}
\usepackage{enumitem}
\usepackage{algorithm}
\usepackage{algpseudocode}
\usepackage[numbers,sort&compress]{natbib}
\usepackage[colorlinks=true,linkcolor=blue!50!black,citecolor=blue!50!black,
            urlcolor=blue!50!black]{hyperref}
\ifanon
  \hypersetup{pdfauthor={}, pdfcreator={LaTeX}, pdfproducer={pdfTeX}}
\else
  \hypersetup{pdfauthor={Sabilashan Ganeshan}, pdfcreator={LaTeX}, pdfproducer={pdfTeX}}
\fi

\theoremstyle{definition}
\newtheorem{definition}{Definition}
\theoremstyle{plain}
\newtheorem{proposition}{Proposition}
\newtheorem{observation}{Observation}

\newcommand{\Llit}{L_{\mathrm{lit}}}
\newcommand{\nid}{n_{\mathrm{id}}}
\newcommand{\nx}{n_{\times}}
\newcommand{\nd}{n_{d}}
\newcommand{\Prec}{\mathcal{H}}

\newcommand{\oeisHol}{2{,}780}          
\newcommand{\oeisRev}{315}              
\newcommand{\oeisRevPct}{11.3\%}        

\title{\bf Where Induction Runs Out: Description-Length Difficulty\\
and the Memorisation Gap in Integer-Sequence Benchmarks}

\ifanon
  \author{Anonymous Author(s)\\Affiliation\\\texttt{email}}
\else
  \author{Sabilashan Ganeshan\\Independent Researcher\\
          \texttt{sabilashanganeshan@gmail.com}}
\fi
\date{}

\begin{document}
\maketitle

\begin{abstract}
Integer sequences from the On-Line Encyclopedia of Integer Sequences (OEIS)
are increasingly used to benchmark mathematical reasoning in language models.
We ask what such benchmarks actually measure, using an exactly computable
reference learner: two-part minimum description length (MDL) over the class of
P-recursive (holonomic) recurrences, evaluated on every prefix of a sequence as
terms arrive.

Three findings follow. First, MDL difficulty is a \emph{parameter count}. The
discovery point $\nd$, the first prefix length at which a symbolic hypothesis
beats verbatim storage, is predicted almost exactly by a combinatorial
identifiability bound on the selected operator's order and degree. It is
invariant to term magnitude: scaling Fibonacci over twelve orders of magnitude
leaves $\nd$ unchanged, because a hypothesis must encode its own initial
conditions and the magnitude cancels.

Second, at scale the learner exhibits a regime our curated corpus could not
produce even once: across $20{,}000$ OEIS sequences, $89.98\%$ of those that
fit a recurrence on some prefix fit \emph{none} at full length. We call this
the \emph{wilderness}---induction acquires a theory, loses it, and never
recovers.

Third, evaluating three language models on sequences stratified by these MDL
regimes \emph{refuted} our pre-registered hypothesis: models do not confabulate
where MDL reports no theory, but hedge appropriately. Confident errors are
\emph{inverted}, concentrating on the easy stratum, where apparent competence
tracks recognition of the sequence rather than induction of its rule.
OEIS-derived benchmarks therefore substantially measure memorisation, and
MDL supplies a cheap, contamination-free difficulty signal they currently lack.
\ifanon
\else
Code, data and all model responses are released at
\url{https://github.com/sabilashang/where-induction-runs-out}.
\fi
\end{abstract}

\section{Introduction}

Integer sequences are an appealing test of mathematical reasoning: the input is
short, the answer is checkable, and the OEIS \citep{oeis} supplies hundreds of
thousands of them for free. They have become a standard benchmark for language
models \citep{dascoli2022,omalley2024}. A benchmark is only as good as its
difficulty measure, yet it is not obvious what makes one sequence harder than
another---growth rate, mathematical depth, or the number of terms needed before
the rule is pinned down.

This paper answers that question with an exactly computable reference learner,
then uses the answer to audit what the benchmark measures. The learner is
two-part MDL \citep{rissanen1978,grunwald2007} over holonomic recurrences,
applied to every prefix of a sequence as terms arrive one at a time. We choose
the class because its MDL hypothesis is exactly computable by linear algebra
over $\mathbb{Q}$ (no search heuristic confounds the measurement), and because
it is the natural symbolic class for integer sequences
\citep{stanley1980,zeilberger1990,kauers2009}.

A natural picture of what such a learner does is dynamic. Given $1, 2, 4, 16$
it entertains $a_n = n$, is refuted, moves to $a_n = 2n$, is refuted, tries
$a_n = 2^{n}$, and perhaps ends up bolting a piecewise branch onto a formula
that no longer fits as a whole. This picture of induction as successive
revision underlies formal learning theory's notion of mind changes
\citep{gold1967,freivalds1993,ambainis1999} and Schmidhuber's account of
interestingness as the derivative of compressibility \citep{schmidhuber2009}.
Yet the modern literature reports only \emph{endpoints}: whether the final
expression is right \citep{dascoli2022}, or how short the final program is
\citep{gauthier2023}. We measure the trajectory, defining the \emph{revision
spectrum} $R(n) = L(n{+}1) - L(n)$ and two exact landmarks, the discovery point
and the stabilisation point.

Our findings, in order:

\begin{enumerate}[leftmargin=*,itemsep=1pt,topsep=3pt]
\item \textbf{Revision is rare on classical sequences, common at scale}
(\S\ref{sec:revfree}). Only $2/61$ classical sequences ever revise, because MDL
is self-regularising: an overfitted recurrence costs more bits than the data it
explains and is never selected. At OEIS scale the rate rises to
\oeisRev/\oeisHol\ (\oeisRevPct), and is length-dependent.

\item \textbf{Difficulty is a parameter count} (\S\ref{sec:idlimited}). The
discovery point is predicted by the identifiability bound
$\nid = (r{+}1)(d{+}1) + s + r$, a combinatorial function of the selected
operator's order and degree alone---exactly in $89\%$ of a labelled classical
corpus and within $\pm1$ in $100\%$, against mean absolute error (MAE) $3.20$
for a compression-based predictor.

\item \textbf{Growth is irrelevant} (\S\ref{sec:growth}). A controlled scaling
experiment over twelve orders of magnitude leaves $\nd$ exactly unchanged,
because magnitude inflates the hypothesis and the literal code at the same rate
and cancels.

\item \textbf{The wilderness} (\S\ref{sec:wilderness}). $89.98\%$ of
sequences that fit some prefix fit nothing at full length---a regime our
classical corpus could not produce once. Structural revisions, when they occur,
are almost always single events rather than cascades.

\item \textbf{Benchmarks measure memorisation} (\S\ref{sec:llm}). Evaluating
three language models on MDL-stratified sequences refutes our pre-registered
hypothesis and yields a stronger one: confident errors concentrate on the
\emph{easy} stratum, where apparent competence turns out to be OEIS
recognition rather than induction.
\end{enumerate}

We regard the last of these as the practical contribution. MDL difficulty is
cheap to compute, independent of term magnitude, and (unlike accuracy on a
public dataset) cannot be inflated by memorisation.

\section{Related work}
\label{sec:related}

\paragraph{Linear complexity profiles.} The closest quantitative relative of
the revision spectrum comes from cryptography, not machine learning. The
linear complexity profile (LCP) $L(s,N)$ of a binary sequence is the length of
the shortest LFSR generating its first $N$ terms, computed by the
Berlekamp--Massey algorithm \citep{massey1969}; its \emph{jumps} as $N$
increases have been studied in depth
\citep{rueppel1986,niederreiter1990,wangmassey1986}. This is precisely a
revision spectrum for the special case of linear recurrences over a finite
field. We differ in three ways that matter. First, our hypothesis class is
holonomic rather than $\mathbb{F}_q$-linear, admitting polynomial coefficients
and hence factorials, binomials and Catalan-type sequences. Second, our metric
is a codelength in bits under a two-part MDL objective rather than a register
length, so hypotheses of different shapes are directly comparable. Third, the
domain and the goal are inverted: the cryptographic literature studies
pseudorandom keystreams and wants profiles to be irregular, whereas we study
structured mathematical sequences and ask what their regularity implies. We
report the LCP as a baseline in Appendix~\ref{sec:lcp}; it separates holonomic
from non-holonomic sequences but cannot distinguish the classes within.

\paragraph{Mind-change complexity.} Formal learning theory quantifies how
often a learner changes its conjecture before converging in the limit
\citep{gold1967,freivalds1993,ambainis1999}. Mind-change complexity is a
worst-case bound (an integer or a constructive ordinal) quantified over an
entire class of targets, and it counts changes. By contrast, the revision
spectrum is an empirical, per-sequence trajectory whose unit is bits. The two
measures are orthogonal in quantification, unit, and object; our result in
\S\ref{sec:revfree} can be read as the observation that the empirical
mind-change count of MDL induction on classical sequences is almost always
exactly one.

\paragraph{MDL stabilization.} \citet{poland2006} prove convergence and loss
bounds for two-part MDL in online prediction over countable model classes, and
give sufficient conditions under which the MDL estimator \emph{stabilizes}.
That work is the theoretical backdrop for \S\ref{sec:revfree}; it establishes that
stabilization eventually happens, whereas we measure, on concrete sequences,
that it happens immediately and characterize exactly when it does not.

\paragraph{Symbolic regression and sequence induction.} \citet{dascoli2022}
train Transformers to infer recurrences for integer and float sequences,
evaluating on an OEIS subset; \citet{gauthier2023} search for short programs
generating OEIS sequences; classical symbolic regression
\citep{schmidt2009,udrescu2020,cranmer2023} and inductive program synthesis
\citep{ellis2021} pursue the same target in other domains. All report final
expressions or final program sizes. Guessing holonomic recurrences from
finitely many terms is standard in computer algebra
\citep{salvy1994,kauers2009,kauers2015}; we use it as an exact oracle rather
than as an end in itself. Compression-based accounts of interestingness
\citep{schmidhuber2009} and of intelligence \citep{chollet2019} motivate
studying the trajectory, but do not compute one.

\section{Setup}

Table~\ref{tab:notation} lists the symbols used below.

\begin{table}[t]
\centering
\small
\begin{tabular}{ll}
\toprule
symbol & meaning \\
\midrule
$n$ & prefix length \\
$r$ & operator order \\
$d$ & degree of the coefficient polynomials \\
$s$ & over-determination slack \\
$c_{ij}$ & integer coefficient of $n^j$ in $p_i$ \\
$L(n)$ & MDL codelength of the prefix of length $n$ \\
$\Llit$ & verbatim (literal) cost of a prefix \\
$L(H)$ & description length of hypothesis $H$ \\
$\rho(n)$ & normalized revision at prefix length $n$ \\
$\nd$ & discovery point \\
$\nid$ & identifiability bound \\
$\nx$ & profitability (compression) threshold \\
$\lambda$ & compression ratio $L(N)/\Llit(N)$ \\
\bottomrule
\end{tabular}
\caption{Notation used throughout.}
\label{tab:notation}
\end{table}

\subsection{Codes}

All description lengths are in bits under prefix-free codes, so the two-part
sum is a genuine codelength and Kraft's inequality holds. For a natural number
$k \ge 0$ we use the Elias gamma code \citep{elias1975},
$\ell(k) = 2\lfloor \log_2 (k{+}1)\rfloor + 1$; a signed integer costs one
further sign bit. The \emph{literal} (verbatim) model of a prefix
$s_{1:n}$ costs
\begin{equation}
\Llit(s_{1:n}) \;=\; \ell(n) + \sum_{i=1}^{n} \ell_{\pm}(s_i).
\end{equation}
This model always applies, which is what makes the MDL codelength finite for
every sequence including those with no recurrence at all.

\subsection{Hypothesis class}

Our class $\Prec$ consists of holonomic operators
\begin{equation}
\sum_{i=0}^{r} p_i(n)\, s_{n+i} \;=\; 0,
\qquad p_i(n) = \sum_{j=0}^{d} c_{ij}\, n^{j},
\label{eq:ansatz}
\end{equation}
with integer coefficients $c_{ij}$, together with the $r$ initial terms needed
to run the recurrence forward. Setting $d = 0$ recovers the C-finite
(constant-coefficient) case. A hypothesis $H \in \Prec$ costs
\begin{equation}
L(H) \;=\; \ell(r) + \ell(d) + \sum_{i,j} \ell_{\pm}(c_{ij})
           + \sum_{k=1}^{r} \ell_{\pm}(s_k).
\label{eq:LH}
\end{equation}
For Fibonacci, $r=2$, $d=0$, the coefficients are $(1,1,-1)$ (encoding
$s_n + s_{n+1} - s_{n+2} = 0$), and the initial terms are $(0,1)$. Then
$\ell(r)=3$, $\ell(d)=1$, each of the three coefficients costs $4$ bits, and
the two initials cost $2$ and $4$ bits, so $L(H)=22$. We admit $H$ only if it
annihilates every available term and its leading polynomial $p_r$ is
nonvanishing on the range used, so that $H$ genuinely determines the sequence
and is therefore a legitimate code for it. Because $H$ reproduces the data
exactly, $L(D\mid H) = 0$ and the two-part objective reduces to
\eqref{eq:LH}.

The MDL codelength of a prefix is then
\begin{equation}
L(n) \;=\; \min\Big\{\; \min_{H \in \Prec,\, H \models s_{1:n}} L(H),
\;\; \Llit(s_{1:n}) \;\Big\}.
\label{eq:Ln}
\end{equation}
Finding the inner minimum is exact: for each $(r,d)$, \eqref{eq:ansatz} is a
homogeneous linear system in the $(r{+}1)(d{+}1)$ unknowns $c_{ij}$, so we
screen for rank deficiency modulo a large prime and then compute the nullspace
exactly over $\mathbb{Q}$, taking the primitive integer vector of each basis
element. We enumerate $(r,d)$ in increasing parameter count with a
branch-and-bound cut, and require the system to be over-determined by a slack
of $s = 2$ equations---the standard guard in computer-algebra guessing
\citep{kauers2009}. \S\ref{sec:revfree} shows this guard is nearly redundant.

\subsection{The revision spectrum}

A revision spectrum records, term by term, whether the next integer was
predicted, merely stored, or forced the learner to change its hypothesis.

\begin{definition}[Revision spectrum]
For a sequence $s$ with $N$ terms, the \emph{raw revision} at $n$ is
$R(n) = L(n{+}1) - L(n)$, and the \emph{normalized revision} is
\begin{equation}
\rho(n) \;=\; \frac{R(n)}{\Llit(s_{1:n+1}) - \Llit(s_{1:n})}
        \;=\; \frac{L(n{+}1) - L(n)}{\ell_{\pm}(s_{n+1})}.
\end{equation}
The \emph{revision spectrum} is the trajectory $\{\rho(n)\}_{n}$.
\end{definition}

The normalization is by the verbatim cost of the newly arrived term, which
makes $\rho$ scale-free and gives it a direct reading: $\rho = 1$ means the
term was absorbed at full verbatim cost (nothing was learned), $\rho = 0$ means
the term was free (the standing hypothesis already implied it), $\rho < 0$
means the term triggered a simplification, and $\rho \gg 1$ means it forced a
costly restructuring.

\begin{observation}[Exactness]
$L(H)$ does not depend on $n$. Hence $R(n) = 0$ \emph{exactly} whenever the
selected hypothesis is unchanged and structural, and every nonzero $R(n)$ is
either a genuine revision or a literal-regime absorption.
\end{observation}

This is why we work with the raw increment rather than with a compression
ratio: a ratio drifts merely because its denominator grows, manufacturing
spurious ``events''.

\begin{definition}[Landmarks]
The \emph{discovery point} $\nd$ is the least $n$ with $L(n) < \Llit(s_{1:n})$,
i.e.\ the first prefix at which structure beats verbatim storage. The
\emph{stabilization point} is the least $n$ after which the selected hypothesis
never changes again. A \emph{structural revision} is a change of hypothesis at
some $n \ge \nd$.
\end{definition}

\subsection{Corpus}

We use $61$ classical integer sequences, each generated from its definition
(never transcribed) and truncated to $N = 34$ terms, labeled by ground-truth
position in the hierarchy: \textsc{poly} (polynomial closed form, $15$),
\textsc{cfin} (C-finite but not polynomial, $16$), \textsc{prec} (holonomic but
not C-finite, $15$), and \textsc{nonh} (provably not holonomic, $15$). The
\textsc{nonh} labels rest on standard results: the primes, the partition and
Bell numbers, the ordered Bell numbers and the classical multiplicative
functions are not holonomic, and the Hofstadter, Kolakoski and Recam\'an
sequences satisfy no linear recurrence with polynomial coefficients.
Table~\ref{tab:corpus} (Appendix~\ref{app:corpus}) summarises the corpus by
class, and OEIS A-numbers are given there for cross-reference
(Figure~\ref{fig:atlas}, Appendix~\ref{app:corpus}).

We deliberately use a corpus we can generate and label rather than a sample of
the OEIS itself, because ground-truth class membership is the independent
variable of every experiment here and the OEIS does not carry it. The pipeline
is class-agnostic and runs unchanged on the OEIS \texttt{stripped} dump; see
\S\ref{sec:limitations}.

\section{Revision is rare, and rarer than it looks}
\label{sec:revfree}

On labelled classical sequences the learner almost never changes its
hypothesis.

Across the corpus, \textbf{$2$ of $61$ sequences} exhibit any structural
revision (the cubes A000578 and the square pyramidal numbers A000330, one
each). Every other sequence either finds its operator once and holds it
forever, or never finds one. The mean number of structural revisions is
$0.033$.

The reason is not the over-determination guard. Sweeping the slack $s$ from
$2$ down to $-2$---that is, permitting systems that are exactly determined or
even under-determined, where a recurrence can interpolate the very data it was
fitted on---barely changes the picture. MDL rejects those hypotheses on its
own: a recurrence with enough free parameters to interpolate $n$ terms costs
more bits under \eqref{eq:LH} than the $\Llit$ of those terms, so
\eqref{eq:Ln} never selects it. Regularization by validation guard and
regularization by codelength are, here, nearly the same constraint, and the
codelength binds first.

\begin{proposition}[informal]
Under \eqref{eq:Ln}, a hypothesis is selected only if it is strictly cheaper
than the data it explains. Since a $\mathbb{Q}$-linear interpolant of $n$ terms
requires $\Theta(n)$ coefficients, each costing $\Omega(1)$ bits, no
interpolating hypothesis is ever selected.
\end{proposition}

The practical consequence is that the churning picture of induction is not a
description of what MDL does. A learner that revises repeatedly is either using
a hypothesis class with free parameters that are cheap relative to the data, or
is not regularizing by description length at all.

\paragraph{At OEIS scale.} The classical corpus is not representative, and the
rate does not survive contact with the full database. Applying the identical
pipeline to $20{,}000$ OEIS sequences \citep{oeis}, \oeisRev\ of \oeisHol\
fully holonomic sequences (\oeisRevPct) at a fixed $30$-term budget
exhibit at least one structural revision---an order of magnitude above the
classical $3.3\%$ (\texttt{oeis\_fixed30.csv}). Two caveats govern this
number and we report both rather than controlling them away.

First, \emph{a sequence counts as holonomic only if an operator is found at
full length}. A recurrence fitted to some prefix but absent at full length is a
distinct phenomenon (\S\ref{sec:wilderness}) and never enters this denominator;
conflating the two inflates the revision rate roughly fivefold.

Second, binning the mixed-length run (\texttt{oeis\_results.csv}; a different population) by available terms gives true-holonomic revision rates
$4.2\%$ ($20$--$22$), $6.1\%$ ($23$--$25$), $9.3\%$ ($26$--$28$),
$5.7\%$ ($29$--$31$) and $12.0\%$ ($32$--$34$).
The dependence is real but modest (far milder than an $n_d$-present
denominator suggested), so any single headline figure remains somewhat
budget-dependent. Cross-study comparisons must fix the term budget. A fully
controlled sweep at fixed length is left to future work for the reasons given
in \S\ref{sec:limitations}.

Structural revisions are overwhelmingly single events rather than cascades: of
sequences that revise at all, the modal and near-universal count is one.

\section{Discovery is identification-limited}
\label{sec:idlimited}

Where a formula is found is a count of free parameters, not a race against
verbatim cost.

If the trajectory is a single jump, the informative statistic is where the jump
occurs. Two thresholds could in principle govern $\nd$:
\begin{align}
\nid &= (r{+}1)(d{+}1) + s + r
 && \text{(identifiability: fewest terms that determine the operator)},\\
\nx  &= \min\{\, n : \Llit(s_{1:n}) > L(H^{*}) \,\}
 && \text{(profitability: fewest terms for which it pays)},
\end{align}
giving the prediction $\nd = \max(\nid, \nx)$. A sequence is
\emph{identification-limited} if $\nid \ge \nx$ and \emph{compression-limited}
otherwise.

Table~\ref{tab:pred} shows the outcome on the $46$ fittable sequences.

\begin{table}[t]
\centering
\small
\begin{tabular}{lrrr}
\toprule
predictor & exact & within $\pm1$ & MAE \\
\midrule
$\max(n_{\rm id}, n_\times)$ & 41/46 & 46/46 & 0.11 \\
$n_{\rm id}$ alone & 41/46 & 46/46 & 0.11 \\
$n_\times$ alone & 0/46 & 0/46 & 3.20 \\
corpus mean & 0/46 & 11/46 & 1.41 \\
\bottomrule
\end{tabular}
\caption{Prediction of the discovery point $\nd$. The identifiability bound
alone accounts for the data; the compression threshold alone does not.}
\label{tab:pred}
\end{table}

The prediction $\max(\nid,\nx)$ is exact for $41/46$ and within $\pm1$ for
$46/46$, with MAE $0.11$. But $\nid$ \emph{alone} achieves identical numbers,
because \textbf{every fittable sequence in the corpus is
identification-limited}: $\nx < \nid$ without exception. Discovery is not
gated by whether the formula pays for itself: by the time the formula can be
pinned down, it already pays.

\paragraph{The dichotomy at scale.} On the OEIS the pooled prediction accuracy
appears to degrade sharply. It does not: it splits. Restricting to sequences
with a full-length operator and conditioning on whether they ever revise gives
Table~\ref{tab:dichotomy}.

\begin{table}[h]
\centering
\small
\begin{tabular}{lrrrr}
\toprule
& $n$ & exact & within $\pm1$ & MAE \\
\midrule
never revises & $2465$ & $91.9\%$ & $99.0\%$ & $0.11$ \\
revises       & $315$  & $0.0\%$  & $48.6\%$ & $4.37$ \\
\bottomrule
\end{tabular}
\caption{Accuracy of the identifiability prediction $\nd=\nid$ on the $2780$
OEIS sequences with a full-length operator (fixed $30$-term budget), split by
whether the sequence ever revises. The pooled figure is dilution, not decay:
the law is near-perfect on non-revisers and fails completely on revisers.}
\label{tab:dichotomy}
\end{table}

Among non-revisers the identifiability law holds as tightly as on the classical
corpus ($91.9\%$ exact; MAE $0.11$, matching to two decimals). Among revisers
it fails completely. Identifiability predicts discovery exactly, with one named
exception---a deceptive prefix, which the revision spectrum itself detects.

\begin{figure}[t]
\centering
\includegraphics[width=\textwidth]{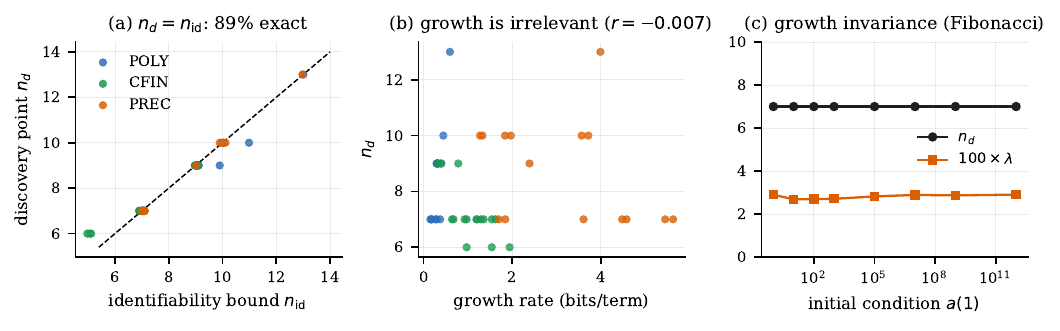}

\caption{(a) The discovery point lies on the identity line against the
identifiability bound. (b) It is uncorrelated with growth rate.
(c) Control A: scaling Fibonacci's initial condition over twelve orders of
magnitude leaves $\nd$ and the compression ratio $\lambda$ exactly unchanged.}
\label{fig:ident}
\end{figure}

Consequently $\nd$ is determined by the pair $(r,d)$ and nothing else. It
correlates with structural size at $r = +0.85$ and with growth rate at
$r = -0.007$ (Figure~\ref{fig:ident}a,b). Mean $\nd$ is $7.19 \pm 0.98$ for
\textsc{cfin}, $8.73 \pm 1.87$ for \textsc{prec} and $8.53 \pm 1.64$ for
\textsc{poly} (one-way analysis of variance (ANOVA) $F = 4.70$, $p = 0.014$); the
separation is
driven mainly by \textsc{cfin}, and \textsc{poly} and \textsc{prec} overlap
substantially. Fittability itself separates the hierarchy perfectly: all $46$
holonomic sequences are fitted and all $15$ non-holonomic ones are not.

\section{Growth invariance}
\label{sec:growth}

Larger terms do not make a sequence easier to discover.

That growth does not matter---already suggested by the near-zero correlation
$r = -0.007$ above---is initially surprising: storing the first $34$
factorials verbatim costs $3519$ bits and storing the first $34$ naturals
costs $315$, so one might expect the expensive sequence to reward a formula
much sooner. It does not: the factorials, the Fibonacci numbers and the
naturals all share the same discovery point $\nd = 7$, though their verbatim
costs differ by an order of magnitude. A controlled experiment isolates the
mechanism.

\paragraph{Control A.} Fix the operator and scale the data. We take Fibonacci
with $a(0)=0$ and $a(1) = k$ for $k = 1, 10, \dots, 10^{12}$. The operator is
identical throughout; only the magnitude of the terms changes. The literal
cost rises from $759$ to $3377$ bits. The discovery point is $\nd = 7$ for
\emph{every} value of $k$, and the final compression ratio
$\lambda = L(N)/\Llit(N)$ is $0.029$ throughout
(Figure~\ref{fig:ident}c).

The reason is visible in \eqref{eq:LH}: a hypothesis must encode its $r$
initial conditions, which are terms of the sequence. Multiplying the sequence
by $10^{12}$ inflates $L(H)$ from $22$ to $98$ bits and $\Llit$ from $759$ to
$3377$ bits---by the same factor. Magnitude information appears on both sides
of the comparison in \eqref{eq:Ln} and cancels. What survives is structure.

\paragraph{Control B.} Vary the operator and the alphabet independently, using
periodic sequences of nominal period $p$ over an alphabet of size $A$. Across
$A \in \{2,3,10,1000\}$ and $p \in \{2,\dots,7\}$ we find
$\nd = \nid = 2p^{*}+3$ exactly in all $24$ cells, where $p^{*}$ is the
\emph{true} period of the generated sequence, with no dependence on $A$.
(In one cell a randomly generated binary block of nominal period $6$ repeated
at period $3$; the law predicted $p^{*}=3$ correctly.) The
compression-limited regime does not appear. This is not an accident of the
corpus: for a C-finite operator of order $r$, $\nid \approx 2r$ while
$\nx \approx L(H)/\bar{\ell} \approx 2r$ as well, since $L(H)$ contains
$r{+}1$ coefficients and $r$ initial terms and $\Llit$ accumulates one
term-cost per step. The two thresholds scale together, and identifiability
wins by the additive slack.

\section{Deceptive prefixes}
\label{sec:deception}

A structural revision occurs when a cheap wrong hypothesis fits a prefix of a
more expensive truth.

\begin{definition}[Deceptive prefix]
A sequence has a \emph{deceptive prefix} of depth $m$ if its first $m$ terms
are consistent with a strictly cheaper hypothesis than the one governing the
whole sequence.
\end{definition}

Deception is the only mechanism we find that produces revisions, and it is
exactly the phenomenon the churning picture imagines. We construct it. Let
\begin{equation}
D_j \;=\; \underbrace{(0\,1)^{j}\,0}_{\text{period } p = 2j+1}
\end{equation}
repeated. Its first $2j$ terms alternate, so a period-$2$ recurrence fits them;
the term at index $2j$ breaks it; the true operator has order $p$ and cannot be
identified until $n = 2p+3$.

\begin{figure}[t]
\centering
\includegraphics[width=0.92\textwidth]{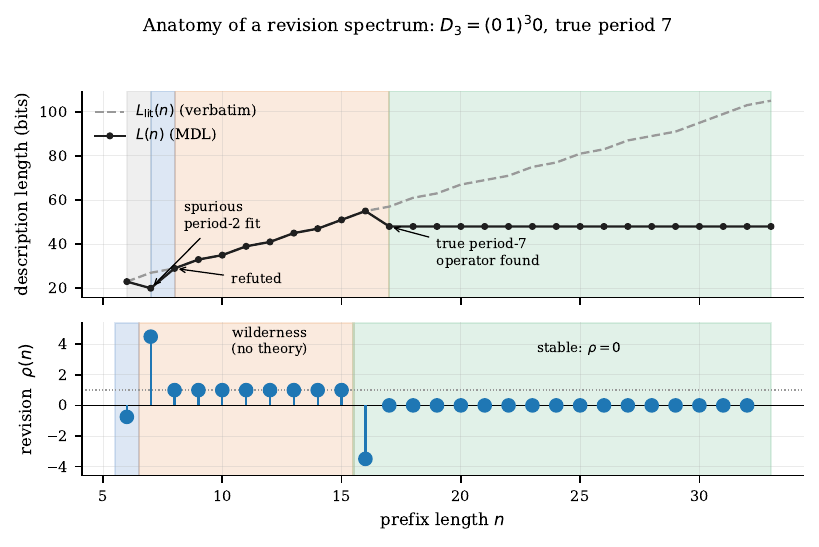}
\caption{The five phases of a revision spectrum on the planted deceptive
sequence $D_3$ (true period $7$). Top: MDL codelength $L(n)$ against the
verbatim cost $\Llit(n)$, shaded by the phase holding at $n$---literal (grey),
spurious period-2 fit (blue), wilderness (orange), stable (green). Bottom: the
normalized revision $\rho(n)$. Since $\rho(n)$ is indexed by its source prefix
length, the revision at $n$ is what installs the state at $n{+}1$; the shading
in the lower panel is shifted one step left accordingly, so each spike is
coloured by the phase it leads into.}
\label{fig:anatomy}
\end{figure}

Figure~\ref{fig:anatomy} shows the resulting spectrum for $D_3$ and reveals a
canonical five-phase shape. Recall that $\rho$ is indexed by its \emph{source}
prefix length, so the revision at $n$ is what installs the state at $n{+}1$:
a literal phase, a \emph{spurious discovery} ($\rho(6) = -0.75$, adopting the
period-2 operator at $n=7$), a \emph{refutation} ($\rho(7) = +4.50$, the
hypothesis is discarded at $n=8$), a \emph{wilderness} in which the learner
holds no theory and absorbs each term at full cost ($\rho = 1$), and a
\emph{true discovery} ($\rho(16) = -3.50$, the true operator is found at
$n = 17$) followed by permanent stability ($\rho \equiv 0$).

\begin{figure}[t]
\centering
\includegraphics[width=\textwidth]{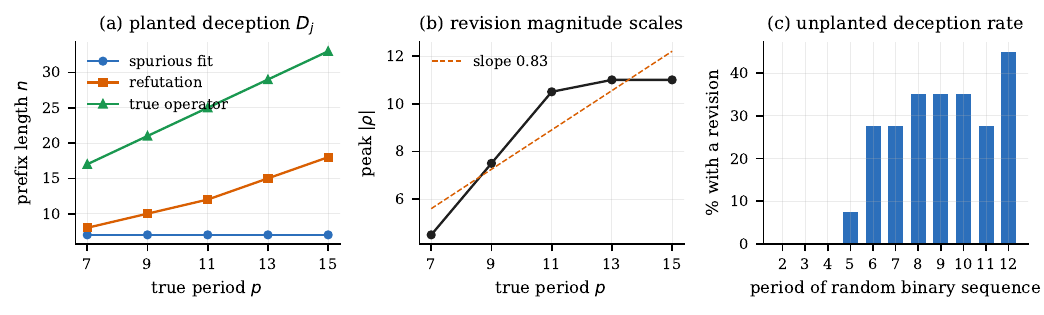}
\caption{Deceptive prefixes in the planted family $D_j$ and in random periodic
sequences. (a) The three landmarks---spurious fit, refutation, and recovery of
the true operator---scale exactly with the true period $p$. (b) Peak revision
magnitude $\max|\rho|$ grows through $p=11$ and then plateaus at $11.0$; the
dashed line is an ordinary least-squares fit over the five planted periods,
drawn only as a straight-line reference against which the plateau is visible,
and its slope is not itself a claim. (c) Fraction of random periodic binary
sequences exhibiting a structural revision, by period.}
\label{fig:deception}
\end{figure}

Sweeping $j$ recovers exact laws (Figure~\ref{fig:deception}). The spurious fit
always appears at $n = 7$; the refutation always at $n = p+1$; the true
operator always at $n = 2p+3$, matching $\nid$ exactly for all
$p \in \{7,9,11,13,15\}$. Peak revision magnitude grows with period up to
$p=11$ ($\max|\rho| = 4.50$, $7.50$, $10.50$ at $p=7,9,11$) and then plateaus
at $11.0$ for $p \ge 13$. We do not have a mechanism for the plateau. Deeper
deception is not merely later, it is \emph{louder}: the size of the spike
measures how much cheaper the refuted theory was than its replacement.

Finally we ask how often deception arises unplanted. Sampling random periodic
binary sequences, the fraction exhibiting a structural revision generally rises
with period: $0\%$ for $p \le 4$, $7.5\%$ at $p=5$, $27.5\%$ at $p=6$,
$35\%$ at $p=10$, $45\%$ at $p=12$, with a dip at $p=11$ ($27.5\%$) and an
overall rate of $21.8\%$ over $440$ sequences. More complex operators leave
more room for a simpler one to fit a prefix by chance.

\section{The wilderness}
\label{sec:wilderness}

Most sequences that fit a recurrence on a short prefix have no recurrence at
full length.

The classical corpus contains sequences that are fitted and sequences that are
never fitted. At OEIS scale a third category dominates: sequences that fit a
recurrence on some prefix and fit \emph{none} at full length. We call this
regime the \emph{wilderness}, after the phase in Figure~\ref{fig:anatomy} in
which the learner holds no theory and absorbs each term at full verbatim cost.

The obvious explanation is truncation. A sequence whose operator requires $40$
terms to identify is indistinguishable, under a $30$-term budget, from one that
has no operator at all. We tested this directly. Taking the $1887$ sequences
exhibiting the pattern, we re-ran the guesser once on every available term;
Table~\ref{tab:wilderness} gives the outcome.

\begin{table}[h]
\centering
\small
\begin{tabular}{lrr}
\toprule
class & count & share \\
\midrule
operator found at full length (truncation artefact) & $29$ & $1.54\%$ \\
no operator, $\ge 40$ terms available (genuine)     & $1698$ & $89.98\%$ \\
no operator, $< 40$ terms (undetermined)            & $160$ & $8.48\%$ \\
\midrule
total & $1887$ & $100\%$ \\
\bottomrule
\end{tabular}
\caption{Truncation audit of the $1887$ OEIS sequences that fit a recurrence on
some prefix but none at the $30$-term budget. Each was re-run on every term the
database holds. Truncation accounts for $1.54\%$; in roughly nine cases in ten
the prefix fit is genuinely not extensible even with $\ge 40$ terms.}
\label{tab:wilderness}
\end{table}

Truncation explains $1.5\%$. The regime is real: in roughly nine cases out of
ten, a recurrence that fits an initial segment is genuinely not extensible, and
the learner is left permanently without a theory despite ample data.

This is the deceptive-prefix mechanism of \S\ref{sec:deception} at scale,
without the second act. In our planted family the learner is deceived, refuted,
and eventually recovers the true operator. On real OEIS data it is deceived,
refuted, and recovers nothing---because most OEIS sequences are simply not
holonomic, while still admitting short-prefix coincidences. Our classical
corpus could not exhibit this at all: its non-holonomic members are so
irregular that no prefix fit is ever cheap enough to be selected. Curating a
corpus for clean class membership removes precisely the phenomenon that
dominates the wild database.

For benchmark construction this is the most consequential regime, and we return
to it in \S\ref{sec:llm}: these are the sequences on which a plausible wrong
answer is available early and no right answer is available at all.

\section{What OEIS benchmarks measure}
\label{sec:llm}

Large language models (LLMs) err confidently on easy sequences and hedge on
hard ones.

\begin{figure}[t]
\centering
\includegraphics[width=\textwidth]{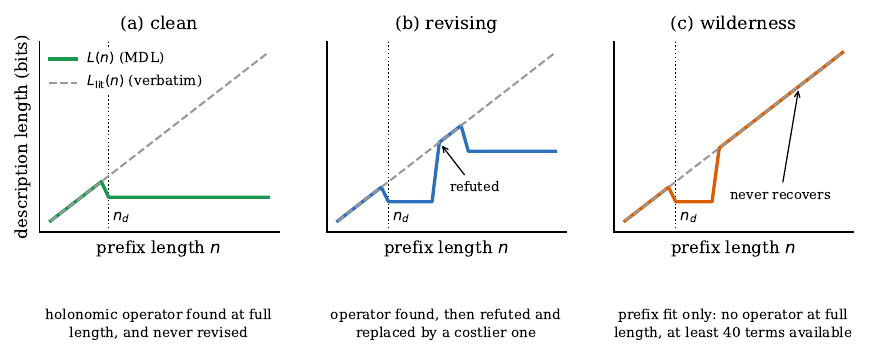}
\caption{The three MDL strata used to stratify the language-model evaluation.
In each panel the solid line is the MDL codelength $L(n)$ and the dashed line
the verbatim cost $\Llit(n)$; the dotted vertical marks the discovery point
$\nd$, where structure first beats verbatim storage. \emph{Clean}: an operator
is found once and held, so $L$ drops at $\nd$ and stays flat. \emph{Revising}:
the operator is later refuted and replaced by a costlier one, so $L$ returns
towards the verbatim cost and then settles at a new, higher level.
\emph{Wilderness}: the prefix fit is refuted and nothing replaces it, so $L$
rejoins $\Llit$ and never leaves it again: the learner is left permanently
without a theory despite ample data. The curves are schematic, drawn to show
the characteristic shapes rather than any measured sequence; the axes are
therefore unticked.}
\label{fig:strata}
\end{figure}

The preceding sections give a difficulty measure that is cheap, exact, and
independent of term magnitude. We now use it to audit the benchmark. We
stratified $60$ sequences into three MDL regimes: \emph{clean} (holonomic, no
revisions), \emph{revising} (holonomic, $\ge 1$ structural revision), and
\emph{wilderness} (prefix fit only, $\ge 40$ terms), sketched in
Figure~\ref{fig:strata}, and evaluated three
LLMs: Claude Sonnet 4.5, GPT-4o and Llama 3.3 70B. Each model saw
the first $20$ terms and was asked, in JSON, for the next three terms, a closed
form or the token \texttt{NO\_FORMULA\_FOUND}, a confidence rating $1$--$5$, and
whether it recognised the sequence. Temperature $0$, one call per
(sequence, model), $180$ calls total; $12$ failed and are reported as failures
rather than imputed. All responses are released.

\paragraph{Our hypothesis was refuted.} We predicted that models would be
confidently wrong in the wilderness: that they would supply a formula where
MDL correctly reports that no theory exists. They do not. All three hedge
appropriately: abstention in the wilderness is $64.7\%$ (Claude, $11/17$),
$95.0\%$ (GPT-4o, $19/20$) and $89.5\%$ (Llama, $17/19$), with mean stated
confidence $3.18$, $2.45$ and $2.68$. Whatever else these models do, they
signal uncertainty when they have no answer.

\paragraph{Confident errors are inverted.} Defining confabulation as a stated
confidence $\ge 4$ together with a wrong continuation, the rate is
\emph{higher} on the easy stratum than the hard one for every model
(Table~\ref{tab:confab}).

\begin{table}[h]
\centering
\small
\begin{tabular}{lrr}
\toprule
model & clean (has formula) & wilderness (none) \\
\midrule
Claude Sonnet 4.5 & $3/17$ \ ($17.6\%$) & $2/17$ \ ($11.8\%$) \\
GPT-4o            & $9/20$ \ ($45.0\%$) & $1/20$ \ ($5.0\%$) \\
Llama 3.3 70B     & $13/20$ \ ($65.0\%$) & $4/19$ \ ($21.1\%$) \\
\bottomrule
\end{tabular}
\caption{Confabulation rate (stated confidence $\ge 4$ together with a wrong
continuation) on the clean stratum, where a closed form exists, against the
wilderness, where none does. For all three models the rate is higher where the
task is easier. Denominators differ because failed calls are reported, not
imputed.}
\label{tab:confab}
\end{table}

\paragraph{Because clean accuracy is recall.} The contamination split explains
the inversion. Conditioning clean-stratum accuracy on whether the model
reported recognising the sequence gives Table~\ref{tab:recognition}.

\begin{table}[h]
\centering
\small
\begin{tabular}{lrrr}
\toprule
model & overall & recognised & not recognised \\
\midrule
Claude Sonnet 4.5 & $12/17$ \ ($70.6\%$) & $8/10$ \ ($80.0\%$) & $4/7$ \ ($57.1\%$) \\
GPT-4o            & $8/20$ \ ($40.0\%$)  & $4/11$ \ ($36.4\%$) & $4/9$ \ ($44.4\%$) \\
Llama 3.3 70B     & $5/20$ \ ($25.0\%$)  & $5/6$ \ ($83.3\%$)  & $0/14$ \ ($0.0\%$) \\
\bottomrule
\end{tabular}
\caption{Clean-stratum exact accuracy split by self-reported recognition.
Llama's performance is entirely recall ($5/6$ on sequences it recognises and
$0/14$ on those it does not), while Claude retains accuracy without recognition
and GPT-4o is essentially unaffected. Recognition is self-reported and may be
unreliable.}
\label{tab:recognition}
\end{table}

\begin{figure}[t]
\centering
\includegraphics[width=\textwidth]{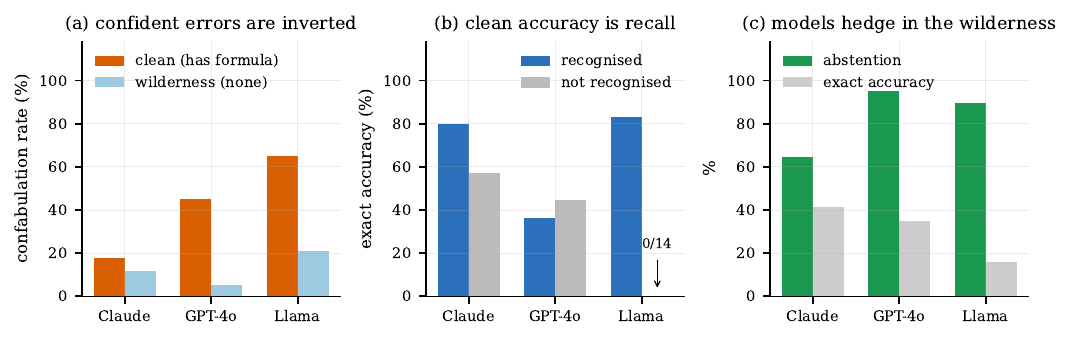}
\caption{LLM behaviour by MDL stratum. (a) Confabulation rate
(stated confidence $\ge 4$ with a wrong continuation) on the clean stratum
(orange) against the wilderness (light blue): confident errors concentrate on
the easy stratum, not the hard one. (b) Clean-stratum exact accuracy split by
whether the model reported recognising the sequence (blue) or not (grey);
Llama scores $0/14$ without recognition. (c) In the wilderness, abstention
(green) against exact accuracy on the next three terms (grey): all three models
hedge, refuting our pre-registered hypothesis. All three panels are percentages
on a fixed $0$--$100$ scale.}
\label{fig:llm}
\end{figure}

Llama is not inferring recurrences; it is retrieving OEIS entries, and it is
most confident exactly where retrieval is available. Claude retains non-trivial accuracy without recognition
($4/7$), and GPT-4o's accuracy is essentially unchanged by recognition, so the
effect varies substantially by model, but the direction of the confabulation
inversion does not (Figure~\ref{fig:llm}).

\paragraph{Reading.} A benchmark whose tractable stratum is answered by
retrieval measures memorisation, and its apparent difficulty gradient partly
tracks how well-known a sequence is rather than how hard it is to induce. The
OEIS is public, indexed, and heavily represented in pretraining corpora, so
this is difficult to avoid by curation alone. What MDL supplies is a difficulty
signal computed from the sequence itself: $\nid$ depends only on operator order
and degree, is unaffected by term magnitude, and cannot be inflated by having
seen the sequence before. We do not claim it measures reasoning; we claim it
measures information-theoretic difficulty, and that the gap between the two is
now itself measurable.

\section{Recommendations for benchmark construction}
\label{sec:implications}

Three concrete suggestions follow from the preceding two sections.

\textbf{Report operator order and degree.} If difficulty is operationalised as
``how many terms must be seen before the rule is inferable'', that quantity is
$\nid = (r{+}1)(d{+}1) + s + r$ under an MDL learner---cheap to compute and
almost perfectly predictive. Strata built on subjective easy/hard labels may be
measuring parameter count, or worse, fame, while appearing to measure reasoning.

\textbf{Report accuracy conditioned on recognition.} A single accuracy figure on
a public dataset cannot distinguish induction from retrieval. Asking the model
whether it recognises the item costs one field and, in our data, changes the
interpretation completely: Llama's clean-stratum accuracy falls from $25\%$ to
$0\%$ once recognised items are removed.

\textbf{Sample the hard regimes deliberately.} Deceptive-prefix sequences, where
a plausible wrong answer is available early, are rare under uniform sampling
($2/61$ in our classical corpus; $21.8\%$ even among random periodic
sequences), and \S\ref{sec:deception} supplies a generator with tunable depth.
Wilderness sequences, where no answer exists at all, test whether a model can
decline---a capability all three models here possess and which no
accuracy-only benchmark rewards.

\section{Limitations}
\label{sec:limitations}

\textbf{Term budget.} The headline OEIS rate uses a fixed $30$-term budget
(\texttt{oeis\_fixed30.csv}); length dependence in \S\ref{sec:revfree} is
from the separate mixed-length run (\texttt{oeis\_results.csv}). A fully
controlled sweep at a fixed $60$-term budget was attempted and abandoned: calibration on a $500$-sequence
slice with instrumented CPU timing gives $1612$ rows/hour wall-clock,
projecting $12.4$ hours per configuration for $20{,}000$ sequences, which
exceeded our compute budget once two encodings were required. An earlier
estimate of $\sim32$ hours was discarded as unreliable. This is the single most
valuable extension of the present work.

\textbf{Corpus.} The $61$-sequence classical corpus is hand-labelled, not
sampled, because ground-truth class membership is the independent variable in
\S\S\ref{sec:idlimited}--\ref{sec:growth}. \S\ref{sec:wilderness} shows the
cost of this directly: curating for clean class membership removes the regime
that dominates the wild database.

\textbf{Hypothesis class.} $\Prec$ is holonomic with $r \le 6$, $d \le 4$. A
richer class---piecewise definitions, general programs, transcendental closed
forms---could show revision where we see none. \S\ref{sec:revfree} is a statement about
MDL over $\Prec$, not about induction in general.

\textbf{Encoding.} $L(H)$ depends on the universal code, and on whether
operators with singular leading coefficients are admitted. The paper reports
the shipped default, \texttt{strict\_leading=False} (singular leading
coefficients allowed, charged via \texttt{singular\_terms}). The stricter
encoding \texttt{strict\_leading=True} is released alongside it in
\texttt{results/ablation\_encoding.csv}. Headline figures are invariant across
the two---$41/46$ exact, MAE $0.11$, two structural revisers, Fibonacci
$\nd = 7$---but several secondary figures move: correlations of $\nd$ with
$L(H^{*})$ and growth, mean $\nd$ and mean $L(H^{*})$ within
\textsc{poly}, reviser identity (cubes/square-pyramidal under the default;
triangular/oblong under the strict encoding), the unplanted deception rate
($21.8\%$ default vs $14.3\%$ under \texttt{strict\_leading=True}; see
\texttt{results/deception\_random\_strict.csv}), and the planted peak-$\rho$
curve. An earlier draft claimed a code change ``could shift $\nx$, though not
$\nid$''; this is too strong, since the selected $(r,d)$ can itself change.

\textbf{A negative result on optimisation.} Skipping the re-solve when the
standing hypothesis still annihilates the new term is \emph{not} sound: under
the default encoding it removes both structural revisions (cubes A000578 and
square pyramidal A000330 fall from one revision each to zero). The cheapest
hypothesis is sometimes not the incumbent even when the incumbent still fits,
so the full re-solve cannot be elided.

\textbf{LLM evaluation.} Sixty sequences across three models is a small sample,
$12/180$ calls failed, and the ``revising'' stratum is under-analysed relative
to the other two. Recognition is self-reported and may be unreliable. The
inverted-confabulation direction is consistent across all three models; the
magnitude is not, and should not be treated as a stable constant.

\section{Conclusion}

We set out to measure the dynamics of symbolic induction, and the measurement
kept correcting us. On a labelled classical corpus the trajectory is a single
jump whose location is a count of operator parameters, invariant to the
magnitude of the data. That clean picture does not survive contact with the
full OEIS: structural revision is an order of magnitude more common, and a
third regime appears that the curated corpus could not produce once. Roughly
nine in ten sequences that fit a recurrence on some prefix fit none at full
length---the learner acquires a theory, loses it, and never recovers.

The same instrument, turned on LLMs, refuted our hypothesis and
replaced it with a better one. Models do not confabulate where no theory
exists; they hedge, and correctly. They confabulate where a theory does exist,
because on that stratum they are retrieving rather than inducing, and retrieval
on a public dataset is confident by construction. The difficulty gradient of an
OEIS benchmark therefore partly tracks fame rather than information content.
Description length offers a difficulty signal that does not: exactly
computable, independent of term magnitude, and immune to having seen the answer
before.

\paragraph{Reproducibility.}
\ifanon
\else
All code, the corpus generator, every model
response, and scripts regenerating each number, table and figure are released at
\url{https://github.com/sabilashang/where-induction-runs-out}.
An archived release is deposited at Zenodo,
\url{https://doi.org/10.5281/zenodo.21830331}.
The classical pipeline runs in under two minutes on a laptop with no GPU.
\fi

\ifanon
\else
\paragraph{Acknowledgements.} AI assistance (Claude, Cursor) was used for
implementation, experiment execution, and drafting. All experimental design
decisions, verification of results, and final claims are the author's.
\fi

\small
\bibliographystyle{plainnat}
\bibliography{refs}

\normalsize
\appendix
\clearpage
\section{Corpus}
\label{app:corpus}

\begin{table}[!ht]
\centering
\small
\begin{tabular}{lrrrrrrr}
\toprule
class & $|C|$ & fitted & mean $n_d$ & range & mean $L(H^*)$ & mean $\lambda_N$ & revisions \\
\midrule
POLY & 15 & 15 & 8.5 & 7--13 & 33 & 0.059 & 2 \\
CFIN & 16 & 16 & 7.2 & 6--9 & 24 & 0.024 & 0 \\
PREC & 15 & 15 & 8.7 & 7--13 & 34 & 0.014 & 0 \\
NONH & 15 & 0 & -- & -- & -- & 1.000 & 0 \\
\midrule
all & 61 & 46 & 8.1 & 6--13 & 30 & 0.270 & 2 \\
\bottomrule
\end{tabular}
\caption{Corpus summary by ground-truth class. $|C|$ is class size, ``fitted''
the number for which a holonomic operator was found, $\lambda_N$ the final
compression ratio, and ``revisions'' the total structural revisions.}
\label{tab:corpus}
\end{table}

\begin{figure}[!ht]
\centering
\includegraphics[width=\textwidth]{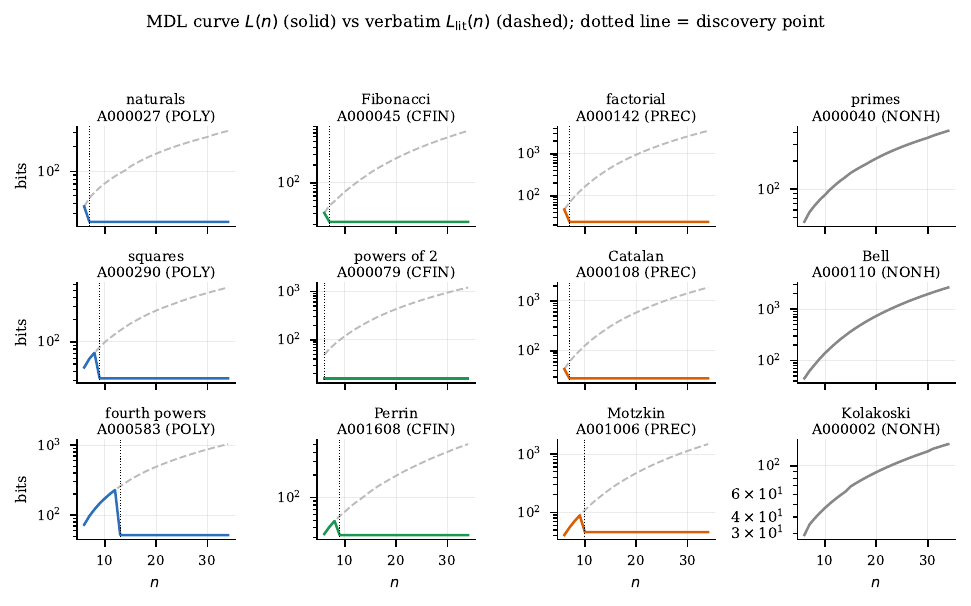}
\caption{MDL curves for representative sequences, three per ground-truth class.
Solid: the MDL codelength $L(n)$. Dashed: the verbatim cost $\Llit(n)$. Dotted
vertical: the discovery point $\nd$. The vertical axis is logarithmic.
Non-holonomic sequences (rightmost column) never separate from the
verbatim cost, which is what having no theory looks like.}
\label{fig:atlas}
\end{figure}

\section{The linear complexity profile baseline}
\label{sec:lcp}

Because the linear complexity profile is the closest existing analogue
(\S\ref{sec:related}), we compute it on the same corpus by reducing each
sequence mod $2$ and running Berlekamp--Massey. The final profile value
separates \textsc{nonh} (mean $11.20$) from the holonomic classes
(means $1.80$, $1.94$, $5.93$) with one-way ANOVA $F = 9.81$, $p < 0.001$.
However it does not distinguish \textsc{poly} from \textsc{cfin} at all
($1.80$ vs $1.94$), because reduction mod $2$ discards precisely the
coefficient and magnitude structure that separates them. The LCP is therefore a
coarse detector of holonomy in this domain, not a substitute for the MDL
spectrum---consistent with its design for binary keystreams rather than
unbounded integer sequences.

\end{document}